\documentclass{ieeeaccess}
\usepackage{cite}
\usepackage{amsmath,amssymb,amsfonts}
\usepackage{algorithmic}
\usepackage{graphicx}
\usepackage{textcomp}
\usepackage[numbers]{natbib}
\def\BibTeX{{\rm B\kern-.05em{\sc i\kern-.025em b}\kern-.08em
    T\kern-.1667em\lower.7ex\hbox{E}\kern-.125emX}}

\usepackage{url}

\begin{document}
\history{Date of publication xxxx 00, 0000, date of current version xxxx 00, 0000.}
\doi{000}

\title{aiSTROM -- A  roadmap for developing a successful AI~strategy}

\author{\uppercase{Dorien Herremans}\authorrefmark{1}, \IEEEmembership{Senior Member, IEEE}}
\address[1]{Singapore University of Technology and Design, Singapore (e-mail: dorien\_herremans@sutd.edu.sg)}
\tfootnote{This work was supported by Singapore Ministry of Education under grant no. MOE2018-T2-2-161, and SRG ISTD 2017 129.\\Preprint accepted for publication in IEEE Access, 2021.}

\markboth
{D. Herremans - aiSTROM}
{D. Herremans - aiSTROM}

\corresp{Corresponding author: Dorien Herremans (e-mail: dorien\_herremans@sutd.edu.sg).}





\begin{abstract} 
  A total of 34\% of AI research and development projects fail or are abandoned, according to a recent survey by Rackspace Technology of 1,870 companies \citep{rackspace2021}. In this perspective paper, a new STrategic ROadMap, aiSTROM, is presented that empowers managers to create an AI strategy. A comprehensive approach is provided that guides managers and lead developers through the various challenges in the implementation process. In the aiSTROM framework, the top $n$ potential projects (typically 3-5) are first identified. For each of those, seven areas of focus are thoroughly analysed. These areas include creating a data strategy that takes into account unique cross-departmental machine learning data requirements, security, and legal requirements. aiSTROM then guides managers to think about how to put together an interdisciplinary artificial intelligence (AI) implementation team given the scarcity of AI talent. Once an AI team strategy has been established, it needs to be positioned within the organization, either cross-departmental or as a separate division. Other considerations include AI as a service (AIaas) and outsourcing development. Looking at new technologies, one has to consider challenges such as bias, the legality of black-box models, and keeping humans in the loop. Next, like any project, value-based key performance indicators (KPIs) need to be defined to track and validate the progress. Depending on the company's risk strategy, a SWOT analysis (strengths, weaknesses, opportunities, and threats) can help further classify the shortlisted projects. Finally, one should make sure that the strategy includes continuous education of employees to enable a culture of adoption. This unique and comprehensive framework offers a practical tool for managers and lead developers.
\end{abstract}


\titlepgskip=-15pt

\maketitle






\section{Introduction}

Artificial Intelligence systems are steadily becoming more and more integrated into our daily lives: from autopilots in our cars \citep{kuutti2020survey}, to personalised suggestions in music streaming app \citep{herremans2020emergence}, and AI diagnostics in the hospital \citep{alloghani2019application}. To survive as a company, it is essential to leverage these technologies to improve one's business processes and offer competitive products or services. But how does one about developing a successful artificial intelligence (AI) strategy? AI systems are unlike traditional software as they have specific requirements and risks \citep{burgess2018industrialising}. According to a study by TechRepublic \citep{TechRepublic2019}, 85\% of AI projects do not manage to bring their intended service to the business. While there are scattered reports and research papers on individual aspects of AI projects, this report aims to bring some of these together from the perspective of how to come to a successful AI project for the business. This paper offers a structured framework to guide an organization's thought processes during the development of an AI strategy, from a multidisciplinary perspective. The proposed framework stems from the experience of the author who has led multiple AI projects in the last decade. 

The emergence of new technologies such as the PyTorch and Keras software, as well as hardware (e.g. GPUs), have made it much easier to develop AI applications \citep{smith2017building}. This evolution has created an immense amount of opportunities for businesses and organizations. When deciding on which opportunities to implement, however, it is important to consider the risks and requirements that come along with such a project, e.g. big data storage, ethical concerns and privacy, to name a few. AI projects are more than just software projects, as they come with their own unique challenges. According to a recent survey of 1,870 companies by \citet{rackspace2021}, 34\% of AI research and development projects fail or are abandoned. As a reason for that failure, a poorly conceived strategy is listed by 31\% of companies.
The proposed aiSTROM roadmap framework will help AI managers to make key decisions to facilitate successful AI projects, by bringing forward the unique challenges that will be faced along the way.

The AI strategic road map tool (aiSTROM) presented in this article offers a concrete roadmap for managers who want to start integrating AI in the organization, as well as developers doing smaller-scale assignments, and consultants overseeing AI projects. In the next section, some background history of AI is introduced, together with why companies need to think about AI strategies now. This is followed by an outline of the aiSTROM framework, which starts by creating a list of potential AI projects, evaluating data strategies, organizing AI development, being aware of technological choices, considering risks and benefits, defining KPIs, and finally, enabling a cultural shift to stimulate AI adoption.


\section{A.I. -- Why now?}
\label{sec:ainow}

The concept of artificial intelligence (AI) is an old one. The first mention of AI is generally attributed to McCarthy in 1956, when he organized the ``Dartmouth Summer Research Project on Artificial Intelligence'', a workshop attended by what turned out to be some of the leading researchers in the field  \citep{buchanan2005very}. Research on AI systems has since advanced from developing rule-based chess computers to deep learning-powered chat agents, and self-driving vehicles. Needless to say, the concept of artificial intelligence has gained immense popularity in the last years \citep{liu2018artificial}.
During the first five days of March 2021, a total of 728 academic papers in the category `Artificial intelligence' were published on the popular computer science preprint repository `arXiv' from Cornell University\footnote{\url{https://arxiv.org/list/cs.AI/current}}. The field of AI has existed for quite a while, yet, it has never been so popular as now. Before we dive into how to tackle AI projects in organizations, we ask the question: what exactly is artificial intelligence?

Some definitions equate artificial intelligence to self-awareness \citep{mitchell2005self}. Looking at the term through a practical, modern-day lens, one notices that the term is used more often to describe models that learn patterns from data and use this to predict future behaviour or instances. Tremendous progress has been made in the last few years in terms of narrow AI systems, i.e. systems with one concrete task: detect if this image is a dog or a cat; or: change the style of this photograph as if it was painted by Monet. The harder task is that reaching a general AI \citep{mccarthy1981some}, or human-level intelligence, whereby the system can solve any new task it is presented with \citep{lieto2018role}. This objective has not yet been reached. In this article, the term AI is used to refer to narrow AI systems that are widely implemented in consumer technology and back-end systems today.

While the first neural network was already proposed in 1944 \citep{fitch1944warren}, its popularity has gone through periods of reduced interest and funding, i.e., the so-called AI winters \citep{gonsalves2019summers}. The first AI winter started in the mid-1970s, due to a combination of circumstances, including disappointing results in machine translation, as well as the Lighthill report, which stated: ``\dots in no part of the field have discoveries made so far produced the major impact that was then promised.'' (James Lighthill, 1973, as quoted by \citep{hendler2008avoiding}). This report was created at the request of the UK's Science Research Council and was meant to evaluate the state of AI's progress. It was not until the 1980s that neural network research became popular again, this time fueled by the work of connectionists such as John Hopfield and David Rumelhart. The last AI major AI winter happened roughly from the late-1980s until the mid-1990s. Its recovery was stimulated by the arrival of the Internet and the availability of large storage devices, both of which led to the age of big data. In addition to this, the current AI boom is further stimulated by the invention of novel, powerful technologies such as convolutional neural networks, along with dedicated processing devices called GPUs. These three factors together, are core contributors to the current AI spring \citep{goodfellow2016deep}, which is offering unprecedented opportunities to businesses.

A report by Price Waterhouse Cooper indicates that  72\% of business leaders think that AI will offer th\'e business advantage of the future \citep{baccala2018ai}. By using AI technologies, one can better understand what happens in an organization down to an unprecedented level, as well as offer novel services and products. \citet{furman2019ai} show that AI technologies are already having a huge impact on the economy. The consultancy agency Accenture estimates that, by 2035, AI will have an added economical value of 8,305 billion USD in the United States alone; 1,079 billion USD in Germany; and 2,068 billion USD in Japan \citep{accenture}. We are thus at the point when AI is reshaping the global economy. Rapid technological disruption is happening through AI technologies, thus empowering companies such as Salesforce, The Trade Desk, and many others. For some companies, however, the adoption of AI is not so straightforward, and it can pose a threat to the continued existence of the organization. Many companies, big and small want to jump on the AI bandwagon, without necessarily knowing what that means to their organization. Gartner \citep{gartner2019} argues that companies should strategize and plan for AI as it can `deliver clever business outcomes'. This calls for an urgent need for companies to implement an AI strategy.
AI has become a driving force in the quick growth of some of the biggest technology companies. Traditional organizations may ask themselves how and when they should start implementing an AI strategy. The general rule of thumb seems to be to start quick, so as to not miss critical opportunities, while newcomers take over the market. A recent report by Gartner also states that these new technologies could lead to a gap between the early adopters and the non- or late-adopters \citep{bughin2018notes}. That being said, one should keep in mind that predictive data systems are not entirely new and they have already provided significant advantages to companies, which can be quickly leveraged. Examples of this include business intelligence systems \citep{alloghani2019application}, customer churn prediction \citep{zhao2005customer}, and APIs for routing \citep{miller2005web}. In the next section, we outline the different elements of the proposed roadmap.

\section{Strategic AI framework (aiSTROM)}

\Figure[h]{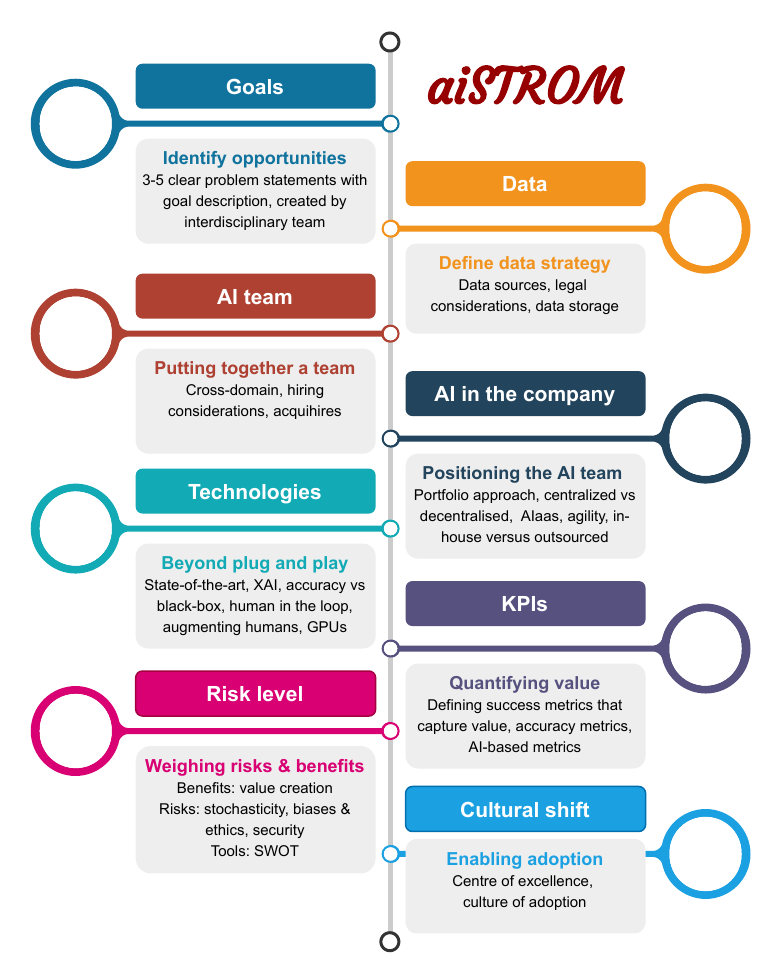}{Strategic roadmap (aiSTROM) for AI implementation.\label{fig:aistrom}}

To guide the strategic decisions when integrating new AI technologies into an organization, this perspective paper offers a roadmap. The aiSTROM framework (see Figure~\ref{fig:aistrom} and described below) aims to provide concrete thinking points and strategic decisions that should not be overlooked when tackling AI projects. Taking some of these items into account will help the organization to successfully select the best AI projects to focus on and give them more insight into the strategic decisions and challenges involved\footnote{The categories below were identified based on the author's extensive domain knowledge as well as a (non-exhaustive) literature search. While all the statements below are backed by references, this paper hence does not offer an exhaustive literature review such as in a meta-review (as categorised by \citet{grant2009typology}), instead, it provides an overview of (categorised) challenges that arise when implementing AI systems. }.

\subsection{Identify opportunities} 
\label{sec:goals}

The myriad of possibilities offered by AI does not come without challenges for organizations: how should they adapt strategically to stay competitive? In other words, which project should the organization focus on first? The aiSTROM framework proposes to first create a list of possible goals and problem statements, each of which can be tackled by an AI system.  This will result in a list of different possible projects with a clear task definition, each with varying levels of risk and different requirements. This list should be created by a team consisting of members with both domain/organization knowledge (e.g. c-level and employees) as well as AI knowledge (e.g. AI consultants or developers). As the organization forms a strategic vision for integrating AI across functional borders in the organization, this will be deeply rooted in the goals that have been outlaid in this step. Having clear goals is an essential step in any successful IT strategy \citep{tallon2000executives, morgan2002business}, it is no different for AI projects.

There are a few things that one should keep in mind when identifying these problem statements. One could identify internal efficiency gaps: where could AI make the organization perform smoother and more efficient? Secondly, what new AI technologies exist and how these could be of potential use to the organization (see Section \ref{sec:tech})? Looking at competitors may also provide some insight into what is possible; what should the organization aim to emulate?; or what is the organization competing with? Are there new business models on the rise that could replace the currently offered services, and should  AI technology be leveraged to adopt these strategies as well? For instance, a traditional taxi company might do well to implement an online booking platform in light of competitors such a Grab and Lift. A similar inspiration for finding AI projects that create value comes from \citet{baccala2018ai}, who state that AI can create value in three different manners: 1) by automating processing; 2) identifying trends from historic data; and 3) by making intelligent predictions that empower human behaviour. Based on current trends one can assume that AI will increasingly form part of business solutions, be it in terms of supporting internal processes or as part of services/products \citep{ransbotham2018artificial}. 

By forming a team that includes both experts with AI expertise, as well as domain experts, the most interesting project ideas can be thought of. Once this list has been made, the top $n$ projects are selected. Here, $n$ could for instance be 3 to 5, unless of course, in the case of a huge AI-based company. When choosing these projects, one most likely will want to maximize the potential impact, given the lowest effort. These top $n$ projects will then be further evaluated and described using the aiSTROM framework laid out in the next subsections. This will result in an analysis that will provide insight into how the implementation of AI projects can be successfully managed, fit into the corporate strategy, as well as ensure that it provides value. By identifying which projects to focus on, time and resources are used efficiently.



\subsection{Data}
\label{sec:data}


We emerged from the last AI winter not just because of the development of new tools (e.g. convolutional neural networks) and fast hardware (i.e., graphic processing units or GPUs), but also due to the prevalence of data \citep{goodfellow2016deep}. Ever since the emergence of the Internet, data is collected everywhere, and the amount of data in warehouses is growing exponentially. Since data is essential for training any successful machine learning algorithm, this is one of the boosting factors of the growing AI industry. \Citet{press6predictions} predicted that the market for big data analytics alone will have over USD 203~billion in revenues worldwide.

The first mention of `big data' can be traced back to \citet{cox1997bigdata}: ``\dots provides an interesting challenge for computer systems: data sets are generally quite large, taxing the capacities of main memory, local disk, and even remote disk. We call this the problem of big data. When data sets do not fit in main memory (in core), or when they do not fit even on local disk, the most common solution is to acquire more resources.''. These days, big data is often characterised by a set of V's. For instance, \citet{russom2011big, laney20013d} define the 3Vs of big data as Volume (it requires a lot of storage space), Velocity (data comes in quickly), Variety (diversity and messiness). These original `Vs' have later been expanded to include Value (how much data contributes to business value) \citep{gantz2012digital},
and even Veracity (biases, noise, and abnormality in data) \citep{white2012digital}. While data is an essential resource for any AI algorithm, it can come with some challenges. Some of the issues that might arise from storing/collecting/using data are discussed below. Considering these issues for each of the above-defined AI problem statements and preparing a proper data strategy is an essential step in the AI framework, or aiSTROM, document. For more details on how to set up a comprehensive data governance strategy, the reader is referred to \citep{tabesh2019implementing}.

According to \citet{rackspace2021}'s survey, the reasons for the failure of AI research and development include a lack of data quality (34\%), and a lack of production-ready data (31\%), thus making data strategy a key consideration in any AI project. In what follows, the focus will be on a few AI-specific considerations related to data strategy.


\subsubsection{Data sources}

A first question that needs to be asked for each of the problem statements is: does the organization already have the necessary data? Maintaining a centralized overview of which data that the organization collects can help the machine learning engineers and designers to ideate which data might be useful for training an AI system. If the organization does not collect data, it is a good idea to start doing so, even before any machine learning models are built. This historical data might prove useful later on for currently undiscovered AI opportunities \citep{davenport2014big}.

\paragraph*{Big data or not? }
The common belief is that the more data is fed to a machine learning algorithm, the better the result. This is partly true, however, it is not always the case that having massive amounts of data results in the best predictive models \citep{junque2013predictive}.  
For instance, having large amounts of `sparse' input features will cause an exponential increase in necessary data and computing resources for obtaining a useful output \citep{brown1995bell}, a phenomenon termed `the curse of dimensionality' by \citet{bellman1961curse}.
In addition, the characteristics of big data cause it to come at quite a cost. Volume necessitates buying or renting large quantities of storage space. While the cost per GB has gone down drastically over the last few years, the cost of \emph{reliable} (secure, backed-up) storage is still higher due to required data replication, continuous administration, and higher hardware requirements.

\paragraph*{Collection versus acquisition}

Collecting data to create a dataset typically requires time. It may also need ethical board approval, quality checks and preprocessing, as well as someone to oversee the process. With this in mind, one might be tempted not to collect all necessary data themselves, but to purchase existing data, provided that it is available to buy. 
Another challenge in data collection is that it can be slow to build a sufficiently large dataset. This process may rely on gathering historic data, such as customer transactions, which may mean waiting until enough transactions have happened. Another type of data collection can be done by annotation, either by a select group of experts or using popular crowdsourcing platforms like Amazon Mechanical Turk \citep{mturk}. In this case, an annotation cost is paid, which is often related to reliability. It is worth keeping in mind that crowdsourcing annotation platforms, even though there are measures to counteract this, often produce noisy, unreliable data. This means that this type of data typically needs more cleaning, preprocessing, and in general, a larger volume of annotations to be useful \citep{wang2015crowdsourcing}. The old saying `garbage in, garbage out', also applies to machine learning models, so the cost of cheap data annotation might in reality be much higher.


Even when a company's AI strategy is not yet set up or finalized, gathering historic data can be beneficial for the (as of yet undefined) future uses. Start collecting as soon as possible seems to be the message \citep{mayer2013big}. Using historic data collected over time can not only be used to train future AI models that have not yet been thought of, but also to facilitate exploratory data analysis. When exploring available data sources, data scientists and domain-experts ask specific questions and try to uncover if the data may support hypotheses that can then be further explored with an AI model. This can be a time-consuming process, but may lead to interesting problem statements or ideas for potential AI systems \citep{azevedo2008kdd}.

\subsubsection{Legal issues: privacy and security}
Through scandals like that of Cambridge Analytics \citep{hinds2020wouldn}, the public is becoming aware that companies are storing their personal data, and may not be keeping it to themselves. Fundamental privacy rights are becoming ever more important. 
Whenever an organization is storing data, legal requirements should be carefully considered so as to keep this data safe and private.

Along with the use of data come certain risks and considerations. While the general notion is that more data enables better AI systems, one has to consider that this can cause privacy issues. By implementing a data sharing policy in the organization, one can reduce the need to create duplicated data and allow engineers come up with ideas and build systems more quickly. The benefits of having data widely available within the organization, are countered by the increased risk of data breaches and privacy issues \citep{figueiredo2017data}. An example of a data breach is the case of Ashley Madison \citep{cross2019media}, a dating website enabling extramarital affairs. Needless to say, their data was highly personal and such breaches should be avoided at all cost. 
If the organization implements greater transparency (of data) between departments to empower AI development, it will require a shift of the traditional organizational boundaries. As an organization, you have to find the balance between quick access to data without asking for permission, so as to enable new ideas and quick prototyping of AI systems, versus adequate security to minimize the risk of data breaches. Even when the internal permissions are secured, one should also protect against malicious external attacks such as exfiltration attacks \citep{d2016data}, in which unauthorized malware or a malicious actor performs an unauthorized data transfer.

Given the above considerations, it is essential that the data requirements necessary to achieve the set-out goals in Step 1 (subsection \ref{sec:goals}) are explored thoroughly and a dedicated internal data privacy manager or external consultant is put in charge of correctly dealing with data given the local privacy laws such as the California Consumer Privacy  Act (CCPA) \citep{de2018guide}, Singapore's Personal Data Protection Act 2012 (PDPA) \citep{wong2017data}  and the European Union's
General Data Protection Regulation (GDPR) \citep{voigt2017eu}. The latter, for instance, grants customers access to data that has been collected about them. It should also be noted that if the organization is active in multiple countries, the privacy laws of those respective countries should all be respected.

\subsubsection{Storing data}

As part of the data governance strategy \citep{tabesh2019implementing}, an organization must also think about where to store data. If a lot of (possibly sensitive) data will be stored, a significant infrastructure investment may be needed along with the manpower needed to administrate the data storage (e.g. data engineer). Instead of organizing internal storage, an external data centre may offer a quick way to get reliable storage with guaranteed security. One should also keep in mind that transferring data is not always fast, hence data should best be stored as close to the user of that data as possible \citep{plank2001managing}. For instance, if the company will be serving many customers in Singapore, but is based in Iceland, it may be worthwhile considering using a Singaporean data storage service.

The data that is collected or purchased may be stored in its raw form in a data lake (unstructured), or it may be preprocessed for indexing into a structured form in a data warehouse (i.e., suited for relational databases such as SQL). Newer types of databases such as MongoDB allow for an in-between type of data, semi-structured, which could, for instance, be an XML file with well-defined fields that may differ between instances \citep{chodorow2013mongodb}. A database specialist should be engaged to make the right decision on which system to implement. This choice should take into account data management principles such as the CAP theorem and PACELC theorem \citep{abadi2012consistency}. The latter refers to the trade-off that different database management systems make between data consistency and data availability in the case of a partition. For more details on the challenges of big data, we refer the reader to \citet{chen2014big}. 


\subsection{The AI team}
When preparing to integrate AI technology into an organization, one should first put together an AI team. In what follows I will discuss some considerations about what the qualities of the AI team may be, as well as possible hiring strategies.

\paragraph*{Hiring considerations} Recently, the demand for technical AI talent has risen starkly \citep{garfinkel2020gartner, cassard2018exponential}. 
Tech giants such as Google and Amazon provide attractive work environments (and compensation) for AI experts. In addition to competing with the tech giants, numerous AI startups are also in the market for hiring from the scarce AI talent pool \citep{talentseer2020}. A survey by \citet{rackspace2021} uncovers that 34\% of failed AI research and development projects are due to a lack of expertise within the organization and 28\% due to a lack of investment in the right people. Before delving into strategies of hiring and keeping AI talent, it is pertinent to first explore what to look for in an employee.

Developing AI systems is technologically challenging as it involves the use of cutting-edge, complex models and architectures. In addition, it necessitates ensuring that the required performance is achieved, both in terms of computing time as well as accuracy, and that the system is scalable. Implementing AI systems includes scientific challenges which require both a domain-knowledge and a technical (AI) skill set. Some of the required skills in an AI development team include \citep{dube2021ai, verma2021investigation}:
\begin{itemize} \setlength{\parskip}{0pt} \setlength{\itemsep}{0pt plus 1pt}
    \item Mathematics and statistics: to properly understand neural network technologies and built novel solutions, a deep understanding of mathematics is essential, often at the PhD level.
    \item Data mining: detecting patterns in large datasets, not just through running a simple model, but by intelligent feature engineering and thinking about what data may be useful. 
    \item Big data engineering: handling and storing data, being aware of privacy and security of the data, scaling up models robustly.
    \item Pattern extraction: it is not just the models that perform pattern extraction, but the developers need to be able to communicate complex technology to less technical supervisors or stakeholders.
    \item Visualisation and user interface (UI) design: so that data can be more easily understood.
    \item Software engineering: developing models in such a way that they are optimized for speed, they are secure, and can scale up. 
    \item Communicating: across a team that has members from a different background.
    \item Continuous learning: new AI technologies are constantly being developed. To stay up to date they have to be curious and motivated.
\end{itemize}

Given all these different skills required for the development of AI systems, there is a need for a diverse, multidisciplinary project team. These days, AI talent is scarce \citep{cbinsights2019} and in high demand \citep{garfinkel2020gartner}. Hence, organizations must come up with a strategy to attract and keep AI experts. 

One possible strategy for obtaining AI talent, other than hiring, is to up-skill existing employees, in technical AI skills as well as vertical AI knowledge, i.e. how to build solutions that target specific customer or business needs \citep{baccala2018ai}. Through internal education, such as workshops and courses, existing employees can be pivoted to become AI experts.
Another strategy is to target recent AI graduates. Due to the scientific challenges involved in developing AI models, PhD level developers may be desired. One way to achieve this is by collaborating closely with academic institutions, or by relocating to strategic places with ample PhD-level AI experts such as the San Francisco Bay area and Boston.

\paragraph*{Acquihire}
There are alternatives to the hiring or up-skilling approach. A possible strategy is that of the acquihire. In an acquihire, a company buys another company, not so much for the product or service they provide, but for the manpower that comes along with the acquisition \citep{varian2020seven, fantasia2016acquihiring}. An example of this is the acquisition of Android by Google in 2005, before any product was developed and when Android still employed less than ten people. The employees had experience designing mobile devices as well as excellent ideas \citep{varian2020seven}. 
In addition, by centering the AI development in the acquired company, the risk of failure can be kept separate from the main business (see also next Section). The strategy of acquihiring has been heavily used by some of the leaders in the AI race: the FAMGA companies (Facebook, Amazon, Microsoft, Google, Apple) \citep{cbinsights2019}.

Once the AI team has been built, an important consideration is to focus on retaining employees through continuous education, benefits, and a desired working environment. In conclusion, the importance of finding trained, capable and diligent manpower for building AI models cannot be overestimated, given the scientific approach that needs to be taken during development.

\subsection{Organizing AI Development} 

Once an AI team has been put together, it is imperative to consider how to position it in the organization. Given the difficulties mentioned above in attracting AI talent, alternatives to in-house development can be considered such as AI as a service (AIaas).

\paragraph*{Positioning the AI team in the organization} 
The management of an AI project can typically be centralized, or decentralized. According to Andrew Ng, a leader in the AI field, a centralized AI team is perfectly positioned to develop an organization-wide AI system \citep{nginterview2019}. To develop an AI application, experts need knowledge about multiple aspects of the organization, and need to understand and access data from many departments. 
In a centralized approach, a chief AI officer might be appointed, who can work across departments with business leaders to drive AI projects forward. 
Centralized teams can take many forms, such as centres of excellence or research labs. The latter often has strong ties to universities. 
In Section \ref{sec:data}, the importance of enabling access to organization-wide data was mentioned. This cross-functionality within the AI team will be crucial for them to develop AI applications by working with all business leaders. Thinking about how to best integrate the AI team in the organization is critical to avoid delay. The decision power of managers and teams will need to be considered, i.e. if any departments need to be involved in decision making, they should be included  in the team. A short reporting line will speed up development \citep{baccala2018ai}. 

Others, like Neil Jacobstein (as per \citet{boulton_2017}) argue for a decentralized approach. In this setup, each department or business team is responsible for developing their own AI systems, supported by the CEO. 
\citet{fountaine2019building} emphasise that there is no one way of organizing the AI team, as it depends on the organization's individual needs. They refer to the AI team in the centralized approach as a `hub', and business units in the decentralized approach as `spokes'. A third, hybrid approach is pointed out by \citet{fountaine2019building}, the `hub-and-spoke' approach, whereby the AI and data analytics capabilities are distributed over both the hub and the spokes. Companies who have done well with this latter approach include Amazon and Google \citep{hub_2020}. In the end, the organization's decision-making structure should be considered. The best approach, given the organization's needs and readiness should be chosen, so as to best allow people to collaborate on unique multidisciplinary projects.



\paragraph*{Portfolio approach}

There are some similarities between AI projects and financial assets. Different projects will entail different levels of risk, due to internal knowledge, the novelty of the project, the adoption readiness of the organization, and data availability. Just like stocks, an organization can implement a portfolio of different AI projects to mitigate risks and maximize potential returns.

This portfolio can be spread across time \citep{bughin2018notes}. In the short term existing, or low-risk technologies (e.g. AIaas) can be implemented to enable quick adoption. In the medium term new, emerging technologies can be explored, together with the value they may bring to the organization. As AI technology might significantly alter the way the organization operates, a culture of adoption as well as openness to change should be ensured. Finally, in the long term, a competitive advantage may be obtained through implementing risky AI projects. 

To mitigate the risk, a portfolio can not only be spread over time, but also across different teams. In this approach, which is reminiscent of venture capital (VC) firms, the (centralized) AI department is given a budget that is divided across its projects or teams. These then have to develop prototypes and proof of concepts, and in each internal funding round, the organization will build upon the projects that have the biggest potential, as well as explore new ones.

\paragraph*{AIaas}

While it is certainly an option to build AI systems within the organization, the adage `do not reinvent the wheel' applies here. Specialized AI companies may have already developed a piece of software that can be useful for the organization. In this case, it is worth considering a Software as a Service (SaaS) or AI as a Service (AIaas) approach \citep{pandl2021drivers}. Especially in the case of non-core technologies, it may not make sense to invest in developing cutting-edge AI models. For instance, Google Maps API used by an insurance company to create a daily schedule of their salesmen. The company does not need to become an expert in routing algorithms, yet it can leverage existing technologies as a service. One thing to keep in mind, is that the organization would be dependant upon potential external price changes of the AIaas. 

\paragraph*{Developing in-house or not? }
Even if the technology does not exist yet, there is an alternative to in-house development. The organization can engage an external development company (i.e., outsource the burden of development and maintenance). According to a recent survey by \citet{rackspace2021}, 41\% of the interviewed companies work with trusted partners/specialist vendors to develop AI applications, whereas 38\% develop them solely in-house.

The decision of whether to develop in-house or not depends on the type of organization, and if it already has the knowledge and manpower in-house. \citet{ransbotham2017reshaping} identify four types of organizations depending on their AI readiness level: pioneers, investigators, experimenters, and passives. These categories have varying levels of understanding and experience with AI. Understanding which category an organization falls into, can help build understanding of the organizational context and thus decide the approach towards implementing AI technologies. While building in-house is a significant commitment, it does ensure that the organization's AI know-how does not fall behind any further. 


Finally, another option is to acquire AI technology is through `Growth by acquisition'. This strategy has the added benefit that it may allow organizations to stay the market cap leader, by buying their competitors. Four companies have been leading AI acquisitions: Google, Apple, Amazon, and Facebook \citep{makridakis2017forthcoming}. 


\paragraph*{Agility}

Looking at implementation strategies, an iterative, agile approach may be best suited for developing AI systems \citep{sharif1999harnessing, hoda2018rise, leijnenagile}. This involves creating prototypes first \citep{shore2007art}, which are then tested and analysed without spending too many resources. These provide stepstones to more polished and well-designed software systems. The insights from initial user tests are then used to reduce risk, and change designs. This reduces the risk overall when scaling up the prototype. It is important to engage the stakeholders early on in the process because what the users may want may differ from the managers' opinion. 

Another recent set of best practices is that of MLOps \citep{alla2021mlops}, which arose from the fusion of machine learning in DevOps practices \citep{karamitsos2020applying}. The practices included in DevOps offer an agile approach to combine an organization's IT development and operations, thus allowing software to be brought into production at a higher velocity. 


\subsection{Technologies}
\label{sec:tech}

Ever since John McCarthy organized the first workshop on AI, i.e. the Dartmouth Summer Research Project on Artificial Intelligence, in the summer of 1956 at Dartmouth College \citep{kaplan2019siri}, numerous machine learning technologies have been developed such as convolutional neural networks, word vector embeddings, reinforcement learning, and transformer networks \citep{goodfellow2016deep, young2018recent}. Giving an overview of these technologies is outside of the scope of this article, hence the reader is referred to \citet{goodfellow2016deep}. The reader should keep in mind though, that even though deep learning technologies are extremely popular, a hybrid approach, that integrates `older' AI technologies such as rules, grammar, and fuzzy logic may be even more efficient \citep{dashtipour2020hybrid}. In what follows, three important considerations when implementing AI techniques are discussed.

\paragraph*{Accuracy versus black-box} An important business consideration regarding AI models is: does the model need to be explainable/comprehensible? This concept is also referred to as XAI \citep{castelvecchi2016can, rai2020explainable}. While deep neural networks have a reputation for being able to reach high performance in terms of prediction accuracy (given enough data), they are not interpretable. This may be an issue if the result of the model needs to be understandable. For instance, a bank doing credit scoring is legally obliged to use understandable rules \citep{waltl2018explainable}. In this scenario, simpler models such as decision trees will have to be used, or rule extraction should be performed on black-box-models \citep{martens2007comprehensible}. The cost of having an explainable model is usually a lower accuracy.

\paragraph*{Human-in-the-loop} To evaluate and measure a model's performance, ground truth labels are typically used. For example, when doing emotion prediction from music, experts need to label music files with emotions \citep{cheuk2020regression}. In a traditional context, supervised learning is used with a split training and test. The former is used to train the model, and the latter to evaluate it. When labelled data is scarce, recent techniques can leverage a small amount of labelled data through semi-supervised or self-supervised learning \citep{lan2019albert}. Other strategies, such as reinforcement learning, put a `human-in-the-loop' and allow new labelled data to be added to the model while it is in production \citep{sutton2018reinforcement}.

\paragraph*{Replacing or augmenting humans?}
Does the organization want the AI system to replace the human previously doing the task, or augment their performance? The latter is a popular approach that aims at creating collaborative/assistive systems instead of autonomous systems, with the intent to make the human's job easier and more effective. For example, one might think that music generation systems can put composers out of business \citep{herremans2017functional}. In reality, those systems can empower composers, to compose quicker, inspire them with new ideas, and let them control generated music on unseen levels, such as through emotion \citep{herremans2017morpheus}.

\paragraph*{Cloud versus in-house hosting}
In Section \ref{sec:ainow}, one of the enablers of the current AI spring was discussed, namely the existence of fast GPUs. GPUs are processing units optimized for the matrix calculations needed when training neural networks. The GPU equipment requirements should be considered for each AI problem statement. Depending on the data size and problem score, a significant number of powerful GPUs may be needed to train the models. There are two broad strategies to follow: 1) renting cloud servers such as Elastic Compute Cloud\footnote{\url{https://aws.amazon.com/ec2}} or Google Compute Engine\footnote{\url{https://cloud.google.com/compute}}, versus 2) buying equipment and running the models in-house. A detailed cost analysis such as that proposed by \citet{emeras2016amazon} can offer valuable insights for making an informed decision. \citet{emeras2016amazon} point out that running the equipment onsite comes with two types of costs: CApital EXpense (capex), and OPerating EXpenses (opex). The former consists of buying machines, storage, interconnecting them, purchasing other room equipment such as cooling, and security. The latter includes the manpower to administrate and operate the machines, energy use, support, and software licences. 
It is no small investment to set up GPU equipment within the organization. Hence, cloud services may offer a quick and cheap path to powerful and scalable GPU power, especially in the early phases of AI projects.

Finally, organizations do not always need the most powerful GPUs in terms of processing power. Recent GPUs, like NVIDIA's Jetson \citep{tang2017enabling}, are ultra-low-power, small devices that can be integrated in Internet of Things (IoT) devices. Depending on the organization's products and use-cases, such devices may be considered to train AI models on small consumer IoT devices.

\subsection{KPIs}


Like any other project, clear success measures need to be defined from the onset, so as to keep track of the project's performance. These clearly defined measures, or Key Performance Indicators (KPIs) will help navigate threats and provide a reporting tool to management in order to justify continued investment in the project. Good success measures are linked to the organization's strategic goals, and may be reviewed often.

Key performance indicators can be used to measure how effectively the organization is reaching its goals, more specifically in this case, the goals influenced by the AI project. It is outside of the scope of this article to describe the entire process of choosing the right KPIs, for that, the interested reader is referred to \citep{parmenter2015key}  and \citep{badawy2016survey}. It is important to point out, however, that an AI implementation might not only affect financial outcomes, but can equally result in value creation for the customer, e.g. better service such as transactional experience \citep{amit2001value}. In the words of A. Einstein as quoted by \citep{juras2020evaluating}: ``Strive not to be a success, but rather to be of value''. Steve Jobs was another proponent of considering value over profit \citep{isaacson2012real}. A similar thought was uttered by Barbara Grosz, Higgins Professor of Natural Sciences at Harvard University, as quoted in \citep{more_2018}: ``it is important to take on the challenge of identifying success measures for AI systems by their impact on people's lives''. As a result, \citet{kiron2019strategy} states that KPIs can centre around customers, costs, processes, or investors. 
Having clear and simple KPIs, gives the development team, and the management concrete objectives to track and aim for.

In a case study of 500 companies who have adopted AI, \citet{wamba2020influence} lists a large number of reported added value through AI such as `Reduce costs by allowing increased asset utilization, as 24/7 executions are possible' in the Transport section; `Save time, improve accuracy and help consumers to find things easily without wasting time' in Sales; `Detect and combat fraud and money laundering' in Banking; `Gain new customers and increase revenue with a real-time analysis solution' in Healthcare; and `Improve the unstructured analysis of data capabilities for the existing solutions' in IT/Telecom. Some of these are value-based in terms of novel services and innovative products. Much of the created value is hidden, however, in the form of more efficient internal processes such as quicker labour time or better customer understanding which can result in better decision making \citep{davenport2018ai}. 
Looking at the KPIs related to AI systems from the survey by \citet{rackspace2021}, it becomes clear that the most two popular KPIs are profit margins (52\%) and revenue growth (51\%). Value-based KPIs, however, are also popular: customer satisfaction/net promoter scores (46\%), process enhancement/improvement (41\%), NPD (New Product Development) (27\%), and time to insight (16\%).


In addition to more traditional KPIs, we should keep in mind that `success' or more concretely `performance accuracy' is not guaranteed for AI systems due to their stochastic nature. While we can have the best intentions to create a model that can identify customers that are most likely to buy  a product, the model might simply not be accurate when looking at new data. Model performance metrics such as accuracy, confusion matrices, or F1-scores give us quick insights into how the model development is advancing and how much potential it will have to be successfully integrated into the business. It is dangerous to just assume that a model will perform well, instead it is better to be open to adapt and willing to pivot in case the data simply does not have any detectable patterns, hence, metrics can provide us with useful insights.

Finally, while KPIs are useful to measure the performance and impact of AI systems, AI can equally be leveraged to calculate better and more insightful KPIs. For instance, Natural Language Processing systems can be used to get real-time feedback on the sentiment of customers \citep{cambria2017sentiment}. Business intelligence systems can be enhanced by AI, and provide easy to interpret metrics that empower decision-makers.

\subsection{Choosing a risk level}

\subsubsection{Risks}

Risk management is essential when managing any project \citep{cooke2002real}. Unlike traditional software, where detailed use case diagrams can describe an entire system in a fairly deterministic and established way \citep{whitten1997systems}, the nature of AI algorithms creates particular kinds of risks and challenges. 

\paragraph*{Stochastic nature}
 When creating AI software, be it a voice assistant, an autonomous driving system or a music album cover generator, the developed models always need to be tested with new, unseen data. A manager can `order' developers to make a model that predicts the customers' next action or to generate music that matches a video, however, there is no guarantee that the model will work accurately. Through several iterations, the development team will need to test how the model performs on different, new datasets. During this process, the model architectures is often adjusted. In addition, since the technology is still rapidly advancing, time needs to be reserved for the continuous education of developers and testing of novel/better technologies that may come out mid-project. Such uncertainties need to be taken into account during development as they can create longer development time, a need for more/better data, research time, and even the need to pivot to a more feasible problem statement.
 Given this uncertainty, it makes sense to develop a prototype first, before investing a lot in an AI idea. Organizations can also look at competitors to see if they have working models. Both approaches can give an idea as to how feasible the desired task is and alert the organization on time if a pivot is needed.

\paragraph*{Biases and ethics}
As machine learning models learn from data, they are only as good as the data they learn from.  The models should therefore be tested for issues such as `learned biases' \citep{dube2021ai}. A famous example of a biased model is the AI recruiting model from Amazon, which was biased against women \citep{yarger2019algorithmic}. Such biases arise because historical datasets are imperfect and contain human biases. In this case, there had been a historical hiring preference for men, which was reflected in the AI predictions. Great care should be taken when building AI models that might inadvertently capture such hidden, unethical biases. These biases are not limited to the field of recruiting. For instance, in healthcare, researchers have established a correlation between wealth,
ethnicity, and health \citep{hague2019benefits}. In \citet{rackspace2021}'s survey, 27\% of failed or discarded AI research and development projects were due to a bias in the algorithm. There is an ethical responsibility that comes with utilizing datasets that contain discrimination based on wealth or ethnicity.

Thoroughly testing AI technologies for biases makes the technology more ethically acceptable. This sort of testing goes beyond looking at loss functions and will require a domain expert or ethics expert to dig up potential problem areas for review and stress testing. There are a few other considerations to make regarding ethics, including ethical dilemmas such as the famous self-driving car example where a vehicle needs to choose between hitting an old lady or a young child. These are its only two options and they have an equal probability. Do you let the algorithm decide, or do you set rules?

\paragraph*{Security} In addition to traditional security issues related to data, a particular kind of security concern that arises only with AI models is related to adversarial attacks \citep{akhtar2018threat}. In such an attack, the user feeds deceptive input to the system to cause a malfunction in the machine learning model and thus fool it. Like many things in AI, the idea came up in movies first, e.g. in Zero History (2010), the character wears a t-shirt that is decorated such that it renders him invisible to surveillance technology. A few years later, in the real world, \citet{xu2020adversarial} designed an `adversarial t-shirt', with up to 74\% success rate. Their shirt introduced `adversarial patches' in the image, that fool person detectors. Adversarial attacks do not just happen on image recognition systems, even simple linear spam classifiers can fall prey to evasion attacks, by spammers who insert `good words' into their spam emails \citep{dalvi2004adversarial}. 
More detailed information and examples on adversarial attacks can be found in \citep{goodfellow2014explaining}.

A separate category of AI systems consists of those used for biometric identification such as face and voice identification systems. They have increased security concerns and will be the target of different types of spoofing algorithms. For instance, voice recognition systems might encounter playback attacks, whereby the perpetrator plays a recorded voice of the rightful user. Spoofing detection AI algorithms try to counter these attacks \citep{balamurali2019toward}. Since these are highly specialized systems, the reader is referred to \citep{dinca2017fall} for more in-depth information.





\paragraph*{Other strategic decisions}

Risk can also result from the chosen strategy, as discussed in the previous sections. Are we using new, hard to implement technologies? Will the system be developed in-house and does the organization have the necessary know-how to do so? If AIaas will be used, is there a price guarantee for the future? Is the AI team integrated into the organization, and do they have access to data across departments? Is a new task being tackled that nobody has tried before, or is it an established task, like face identification? Finally, for each identified AI system that the organization wants to implement, the trade-off between risk and benefits can be plotted in a graph for each potential project.


\subsubsection{Benefits}

Any digital business strategy should keep value-creation in mind \citep{bharadwaj2013digital}. 
The aim of developing any AI system should be to create value for either the customer or internally for the business. In this phase, the organization should explore what value will be created by each of the AI projects. Examples of added value include a more secure banking experience, personalized shopping recommendations, and health alerts based on biometric feedback from a smartwatch. Ideally, these are captured by the KPIs, but they may be broader and harder to measure. 


\subsubsection{SWOT}

Just like when weighing financial investments, one can opt to implement high-risk, high-value AI models (e.g. highly innovative services and products), or choose to stick with low-risk, low-value models (e.g. automatic internal repetitive processes). Alternatively, some technologies offer a blend of the two. High-value, low-risk models include those that automatise complicated mathematical decision support systems; and low-value, high-risk include those that try to predict complex customer behaviour.

Depending on the risk strategy of the organization, a portfolio of different models may be adopted to have a spread of risk, as discussed above in the portfolio approach. The information gathered in the aiSTROM roadmap framework can be used in a Strengths/weaknesses/opportunities/threats (SWOT) analysis \citep{gurel2017swot} to assist managers to create an informed decision and put together a balanced portfolio. The risk profile of the organization will also affect its implementation strategy as discussed below.

\subsection{Enabling a cultural shift through education}

The employees directly involved with the AI implementation, either as managers or developers, should be knowledgeable of AI technologies. The technical experts in the AI team can also be part of the team that ensures the continuous education of not only the employees involved in the project but the rest of the organization as well. Other employees, as far removed from the AI project as they might be, still contribute to the culture of the organization and thus help form the vision of the products or services, and they may even be the user of the AI system if it is related to internal processes.

One way to achieve this internal education is by setting up a centre of excellence within the organization, which is responsible for creating awareness and knowledge of AI technologies within the organization \citep{davenport2019set}. According to Wilson Pang, Chief Technology Officer of Appen \citep{pang_2020}, these centres can be seen as think thanks, run by experts in AI. Having an AI centre of excellence can help enable a shift of the organization's culture to be inclusive of AI thinking, and hence let people embrace AI. When employees are educated, they can be involved in brainstorming workshops to help think about how to embed AI technologies in the organization. Through this, the organization is not just leveraging AI for competitive advantage, but embracing these technologies as a driving force of the organization, and letting go of legacy systems. Through education, employees can be empowered to be part of the AI strategy of the organization.

In addition, when AI literacy is stimulated among employees, they may replace their potential fears of being replaced, by the knowledge that AI can empower them (as discussed in Section \ref{sec:tech}), and how they can work in conjunction with the system in a symbiotic way. This promotes a culture of adoption. \citet{fountaine2019building} point out that `only 8\% of firms are engaging in core practices that support widespread adoption', which leads to a failure when the organization tries to scale up AI technologies. Creating an AI culture that supports broad organization-wide adoption is essential.


Finally, it is worth noting that employee training is not a one-time thing. AI technologies constantly change, and hence there should be a culture of continuous education. This can work beneficial towards keeping employees engaged with the AI integration in the organization and help facilitate a cultural shift of AI adoption. 




\section{Conclusion}

This perspective paper presents aiSTROM, a strategic framework that guides managers through the process of integrating AI in the organization (Figure~\ref{fig:aistrom}). aiSTROM provides managers with a framework to think strategically about the challenges involved in AI projects. The first step in the proposed framework is identifying a top $n$ list of potential problem statements/tasks that can be tackled with AI. In the following steps, the potential projects on this list will be analysed according to different factors. 

Data is the first important factor as it is the throbbing heart of any machine learning system. Strategic issues that are important to keep in mind when selecting data sources are discussed, together with legal issues associated with them, and options related to secure storage. Secondly, due to the scarcity of AI talent, important challenges and options related to composing an AI team are examined. In addition, given the interdisciplinary of AI projects, some consideration is given to where to position the team in the organization, and if the organization should consider alternative solutions such as AIaas and acquisitions. 

In the aiSTROM framework, several considerations related to technology are taken into account. These include black-box versus explainable models, human-in-the-loop models, replacing versus empowering human users, and finally, issues related to GPU equipment. 
Like any project, AI projects also need proper, value-based KPI metrics to help measure and justify their progress to management.
Finally, risks and benefits that need to be considered when performing a SWOT analysis are discussed, and a case is made for the importance of enabling a cultural shift within the organization to support widespread adoption.

Armed with a thorough analysis of the potential AI projects resulting from aiSTROM, managers can make informed decisions, that are bound to increase the success of the organization's AI strategy. 

In future research, the proposed aiSTROM framework should be validated in real-world situations, so as to go beyond relying on data and conclusions drawn from existing surveys and literature.




\small

\bibliographystyle{IEEEtranN}
\bibliography{paper, mybibfile}

\begin{IEEEbiography}[{\includegraphics[width=1in,height=1.25in,clip,keepaspectratio]{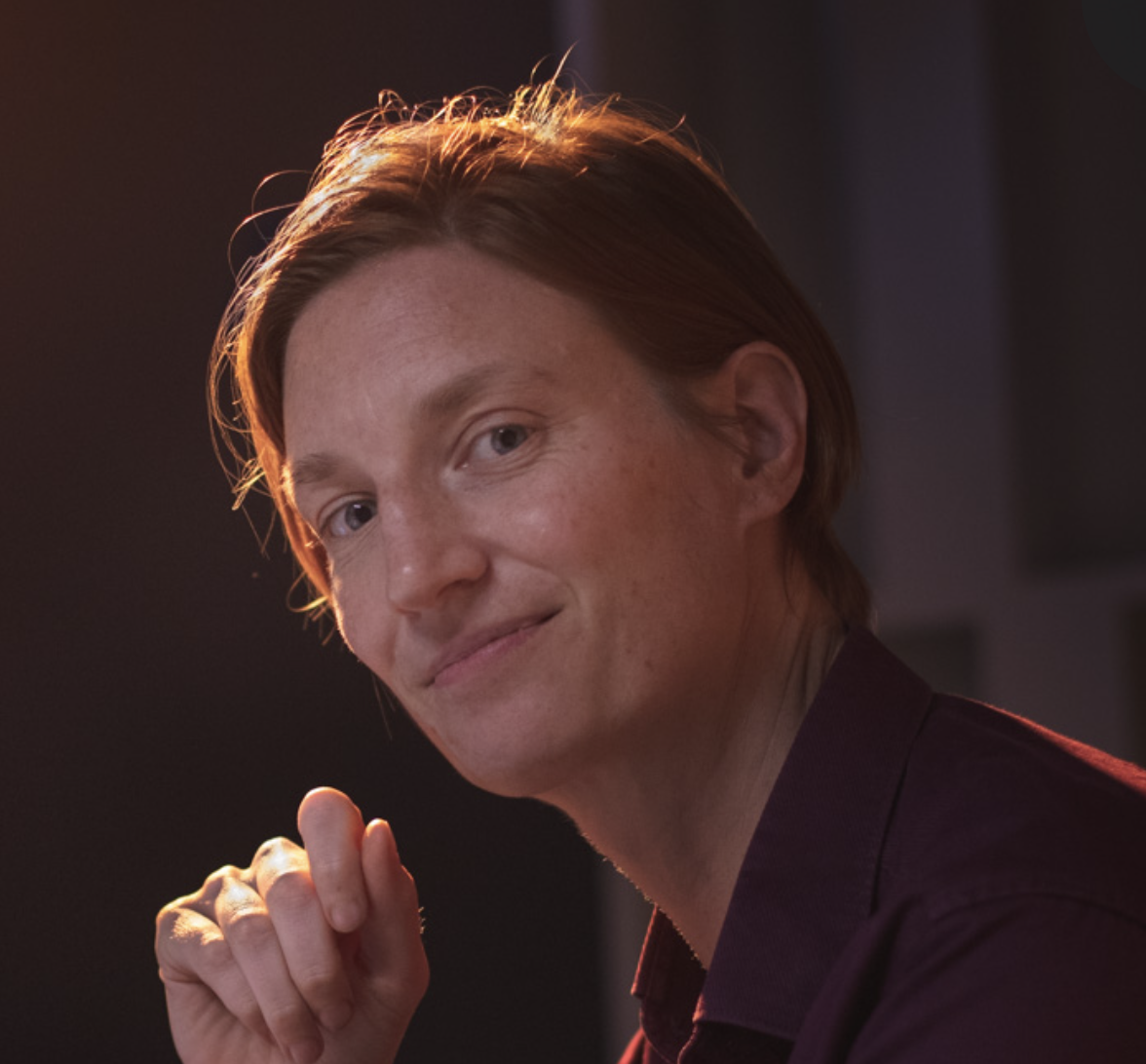}}]{Dorien Herremans} (M'13--SM'17) received her Ph.D. degree in Applied Economics from the University of Antwerp. She was awarded a Marie-Curie fellowship to work at the Centre for Digital Music, Queen Mary University of London. Before that, she graduated as a Business Engineer in Management Information Systems at the University of Antwerp in 2005, after which she worked as a Drupal Consultant and as an IT Lecturer at the world's leading hospitality business school, Les Roches, in Bluche, Switzerland. Currently she is an Assistant Professor with the Singapore University of Technology and Design. At SUTD she is also Director of SUTD Game Lab and heads both the AMAAI (Audio, Music, and AI) lab as well as the AIFi (AI for finance) group. Strategic thinking as well as novel applications of AI technologies are her passion. She serves on numerous committees and boards and has given invited talks around the world. Dr. Herremans is on the Singapore 100 Women in Tech 2021 list, which recognises and celebrates women based in Singapore who have been inspiring and have made significant contributions to the tech industry.

\end{IEEEbiography}

\EOD



\end{document}